# Combining Hand-crafted Rules and Unsupervised Learning in Constraint-based Morphological Disambiguation

Kemal Oflazer and Gökhan Tür
Department of Computer Engineering and Information Science
Bilkent University, Bilkent, Ankara, TR-06533, TURKEY
{ko,tur}@cs.bilkent.edu.tr

## Abstract

This paper presents a constraint-based morphological disambiguation approach that is applicable languages with complex morphology–specifically agglutinative languages with productive inflectional and derivational morphological phenomena. In certain respects, our approach has been motivated by Brill's recent work (Brill, 1995b), but with the observation that his transformational approach is not directly applicable to languages like Turkish. Our system combines corpus independent hand-crafted constraint rules, constraint rules that are learned via unsupervised learning from a training corpus, and additional statistical information from the corpus to be morphologically disambiguated. The hand-crafted rules are linguistically motivated and tuned to improve precision without sacrificing recall. The unsupervised learning process produces two sets of rules: (i) *choose rules* which choose morphological parses of a lexical item satisfying constraint effectively discarding other parses, and (ii) *delete rules*, which delete parses satisfying a constraint. Our approach also uses a novel approach to unknown word processing by employing a secondary morphological processor which recovers any relevant inflectional and derivational information from a lexical item whose root is unknown. With this approach, well below 1% of the tokens remains as unknown in the texts we have experimented with. Our results indicate that by combining these hand-crafted, statistical and learned information sources, we can attain a recall of 96 to 97% with a corresponding precision of 93 to 94%, and ambiguity of 1.02 to 1.03 parses per token.

## 1 Introduction

Automatic morphological disambiguation is a very crucial component in higher level analysis of natural language text corpora. Morphological disambiguation facilitates parsing, essentially by performing a certain amount of ambiguity resolution using relatively cheaper methods (e.g., Güngördü and Oflazer (1995)). There has been a large number of studies in tagging and morphological disambiguation using various techniques. Part-of-speech tagging systems have used either a statistical approach where a large corpora has been used to train a probabilistic model which then has been used to tag new text, assigning the most likely tag for a given word in a given context (e.g., Church (1988), Cutting et al. (1992), DeRose (1988)). Another approach is the rule-based or constraint-based approach, recently most prominently exemplified by the Constraint Grammar work (Karlsson et al., 1995; Voutilainen, 1995b; Voutilainen et al., 1992; Voutilainen and Tapanainen, 1993), where a large number of hand-crafted linguistic constraints are used to eliminate impossible tags or morphological parses for a given word in a given context. Brill (1992; 1994; 1995a) has presented a transformation-based learning approach, which induces rules from tagged corpora. Recently he has extended this work so that learning can proceed in an unsupervised manner using an untagged corpus (Brill, 1995b). Levinger et al. (1995) have recently reported on an approach that learns morpholexical probabilities from untagged corpus and have the used the resulting information in morphological disambiguation in Hebrew.

In contrast to languages like English, for which there is a very small number of possible word forms with a given root word, and a small number of tags associated with a given lexical form, languages like Turkish or Finnish with very productive agglutinative morphology where it is possible to produce thousands of forms (or even millions (Hankamer, 1989)) for a given root word, pose a challenging problem for morphological disambiguation. In English, for example, a word such as *make* or *set* can be verb or a noun. In Turkish, even though there are ambiguities of such sort, the agglutinative nature of the language usually helps resolution of such ambiguities due to restrictions on morphotactics. On the other hand, this very nature introduces another kind of ambiguity, where a lexical form can be morphologically interpreted in many ways, some with totally unrelated roots and morphological features, as will be exemplified in the next section.

Our previous approach to tagging and morphological disambiguation for Turkish text had employed a constraint-based approach (Oflazer and Kuruöz, 1994) along the general lines of similar previous work for English (Karlsson et al., 1995; Voutilainen et al., 1992; Voutilainen and Tapanainen, 1993). Although the results obtained there were reasonable, the fact that all constraint rules were hand crafted, posed a rather serious impediment to the generality and improvement of the system.

In this paper we present a constraint-based morphological disambiguation approach that uses unsupervised learning component to discover some of the constraints it uses in conjunction with hand-crafted rules. It is specifically applicable to languages with productive inflectional and derivational morphological processes, such as Turkish, where morphological ambiguity has a rather different nature than that found in languages like English. Our approach starts with a set of corpus-independent hand-crafted rules that reduce morphological ambiguity (hence improve precision) without sacrificing recall. It then uses an untagged training corpus in which all lexical items have been annotated with all possible morphological analyses, incrementally proposing and evaluating additional (possibly corpus dependent) constraints for disambiguation of morphological parses using the constraints imposed by unambiguous contexts. These rules choose or delete parses with specified features. In certain respects, our approach has been motivated by Brill's recent work (Brill, 1995b), but with the observation that his transformational approach is not directly applicable to languages like Turkish, where tags associated with forms are not predictable in advance.

In the following sections, we present an overview of the morphological disambiguation problem, highlighted with examples from Turkish. We then present the details of our approach and results. We finally conclude after a discussion and evaluation of our results.

## 2 Tagging and Morphological Disambiguation

In almost all languages, words are usually ambiguous in their parts-of-speech or other lexical features, and may represent lexical items of different syntactic categories, or morphological structures depending on the syntactic and semantic context. Part-of-speech (POS) tagging involves assigning every word its proper part-of-speech based upon the context the word appears in. In English, for example a word such as *set* can be a verb in certain contexts (e.g., He *set* the table for dinner) and a noun in some others (e.g., We are now facing a whole *set* of problems).

In Turkish, there are ambiguities of the sort above. However, the agglutinative nature of the language usually helps resolution of such ambiguities due to the restrictions on morphotactics. On the other hand, this very nature introduces another kind of ambiguity, where a whole lexical form can be morphologically interpreted in many ways not predictable in advance. For instance, our full-scale morphological analyzer for Turkish returns the following set of parses for the word *oysa*:[1,2]

1. [[CAT CONN] [ROOT oysa]]
   (on the other hand)

2. [[CAT NOUN] [ROOT oy] [AGR 3SG]
   [POSS NONE] [CASE NOM]
   [CONV VERB NONE]
   [TAM1 COND] [AGR 3SG]]
   (if it is a vote)

3. [[CAT PRONOUN] [ROOT o] [TYPE DEMONS]
   [AGR 3SG] [POSS NONE]
   [CASE NOM] [CONV VERB NONE]
   [TAM1 COND][AGR 3SG]]
   (if it is)

4. [[CAT PRONOUN] [ROOT o] [TYPE PERSONAL]
   [AGR 3SG] [POSS NONE] [CASE NOM]
   [CONV VERB NONE] [TAM1 COND]
   [AGR 3SG]]
   (if s/he is)

5. [[CAT VERB] [ROOT oy] [SENSE POS]
   [TAM1 DES] [AGR 3SG]]
   (wish s/he would carve)

On the other hand, the form *oya* gives rise to the following parses:

1. [[CAT NOUN] [ROOT oya] [AGR 3SG]
   [POSS NONE] [CASE NOM]] (lace)
2. [[CAT NOUN] [ROOT oy]  [AGR 3SG]
   [POSS NONE] [CASE DAT]] (to the vote)
3. [[CAT VERB] [ROOT oy]  [SENSE POS]
   [TAM1 OPT] [AGR 3SG]]  (let him carve)

and the form *oyun* gives rise to the following parses:

---

[1] Output of the morphological analyzer is edited for clarity, and English glosses have been given.

[2] Glosses are given as linear feature value sequences corresponding to the morphemes (which are not shown). The feature names are as follows: CAT-major category, TYPE-minor category, ROOT-main root form, AGR -number and person agreement, POSS - possessive agreement, CASE - surface case, CONV - conversion to the category following with a certain suffix indicated by the argument after that, TAM1-tense, aspect, mood marker 1, SENSE-verbal polarity, DES- desire mood, IMP-imperative mood, OPT-optative mood, COND-Conditional

```
1. [[CAT NOUN] [ROOT oyun] [AGR 3SG]
       [POSS NONE] [CASE NOM]] (game)

2. [[CAT NOUN] [ROOT oy]   [AGR 3SG]
       [POSS NONE] [CASE GEN]] (of the vote)
3. [[CAT NOUN] [ROOT oy]   [AGR 3SG]
       [POSS 2SG] [CASE NOM]]   (your vote)

4. [[CAT VERB] [ROOT oy]   [SENSE POS]
       [TAM1 IMP] [AGR 2PL]] (carve it!)
```

On the other hand, the local syntactic context may help reduce some of the ambiguity above, as in:[3]

| sen-in | oy-un .. |
| PRON(you)+GEN | NOUN(vote)+POSS-2SG |
| your | vote |

| oy-un | reng-i .. |
| NOUN(vote)+GEN | NOUN(color)+POSS-3SG |
| (NOUN-GEN | NOUN-POSS form) |
| color | of the vote |

| oyun | reng-i .. |
| NOUN(game) | NOUN(color)+POSS-3SG |
| game | color |
| (NOUN | NOUN-POSS form) |

using some very basic noun phrase agreement constraints in Turkish. Obviously in other similar cases, it may be possible to resolve the ambiguity completely.

There are also numerous other examples of word forms where productive derivational processes come into play:[4]

```
geldiGimdeki (at the time I came)

[[CAT VERB] [ROOT gel] [SENSE POS]
  (basic form)
[CONV NOUN DIK] [AGR 3SG]
[POSS 1SG] [CASE LOC]
  (participle form)
[CONV ADJ REL]]
(final adjectivalization by the
 relative (ki) suffix)
```

Here, the original root is verbal but the final part-of-speech is adjectival. In general, the ambiguities of the forms that come before such a form in text can be resolved with respect to its original (or intermediate) parts-of-speech (and inflectional features), while the ambiguities of the forms that follow can be resolved based on its final part-of-speech.

The main intent of our system is to achieve a *morphological ambiguity reduction* in the text by choosing for a given ambiguous token, a subset of its parses which are not disallowed by the syntactic context it appears in. It is certainly possible that a given token may have multiple correct parses, usually with the same inflectional features or with inflectional features not ruled out by the syntactic context. These can only be disambiguated usually on semantic or discourse constraint grounds.[5]

We consider a token *fully disambiguated* if it has only one morphological parse remaining after automatic disambiguation. We consider as token as correctly disambiguated, if one of the parses remaining for that token is the *correct* intended parse.[6] We evaluate the resulting disambiguated text by a number of metrics defined as follows (Voutilainen, 1995a):

$$Ambiguity = \frac{\#Parses}{\#Tokens}$$

$$Recall = \frac{\#Tokens\ Correctly\ Disambiguated}{\#Tokens}$$

$$Precision = \frac{\#Tokens\ Correctly\ Disambiguated}{\#Parses}$$

In the ideal case where each token is uniquely and correctly disambiguated with the correct parse, both recall and precision will be 1.0. On the other hand, a text where each token is annotated with all possible parses,[7] the recall will be 1.0 but the precision will be low. The goal is to have both recall and precision as high as possible.

## 3 Constraint-based Morphological Disambiguation

This section outlines our approach to constraint-based morphological disambiguation incorporating unsupervised learning component. Our system with the structure presented in Figure 1 has three main components:

1. the preprocessor,
2. the learning module, and
3. the morphological disambiguation module.

Preprocessing is common to both the learning and the morphological disambiguation modules. The module takes as input to the system raw Turkish text and preprocesses it in a manner to be described shortly.

If the text is to be used for training, the learning module then

1. applies an initial set of linguistically motivated hand-crafted constraint rules to choose and/or delete certain parses, and

---

[3]With a slightly different but nevertheless common glossing convention.

[4]Upper cases in morphological output indicates one of the non-ASCII special Turkish characters: e.g., G denotes ğ, U denotes ü, etc.

[5]For instance the third and fourth parses for *oysa* above.

[6]It is certainly possible that, a parse that is deleted may also be a valid parse in that context.

[7]Assuming no unknown words.

2. uses an unsupervised learning procedure to induce some additional (an possibly corpus dependent) rules to choose and delete some parses.

Morphological disambiguation of previously unseen text proceeds as follows:

1. The hand-crafted rules are applied first.
2. Certain parses are deleted using context statistics on the corpus to be tagged.
3. Rules learned to choose and delete parses are then applied.

### 3.1 The Preprocessor

The preprocessing module takes as input a Turkish text, segments it into sentences using various heuristics about punctuation, tokenizes and runs it through a wide-coverage high-performance morphological analyzer developed using two-level morphology tools by Xerox (Karttunen, 1993). This module also performs a number of additional functions:

- it groups *lexicalized collocations* such as idiomatic forms, semantically coalesced forms such as proper noun groups, certain numeric forms, etc.

- it groups any *compound verb formations* which are formed by a lexically adjacent, direct or oblique object, and a verb, which for the purposes of syntactic analysis, may be considered as single lexical item: e.g., *saygı durmak* (to pay respect), *kafayı yemek* (literally *to eat the head* – to get mentally deranged), etc.

- it groups *non-lexicalized collocations*: Turkish abounds with various non-lexicalized collocations where the sentential role of the collocation has (almost) nothing to do with the parts-of-speech of the individual forms involved. Almost all of these collocations involve duplications, and have forms like $\omega + x \; \omega + y$ where $\omega$ is the duplicated string comprising the root and certain sequence of suffixes and $x$ and $y$ are possibly different (or empty) sequences of other suffixes.

The following is a list of multi-word constructs for Turkish that we handle in our preprocessor. This list is not meant to be comprehensive, and new construct specifications can easily be added. It is conceivable that such a functionality can be used in almost any language. (See Oflazer and Kuruöz (1994) and Kuruöz (1994) for details of all other forms for Turkish.)

1. duplicated optative and 3SG verbal forms functioning as manner adverb. An example is *koşa koşa*, where each lexical item has the morphological parse

   ```
   [[CAT VERB] [ROOT koS] [SENSE POS]
       [TAM1 OPT] [AGR3SG]]
   ```

   The preprocessor recognizes this and generates the feature sequence:

   ```
   [[CAT VERB] [ROOT koS] [SENSE POS]
       [TAM1 OPT] [AGR 3SG]
       [CONV ADVERB DUP1] [TYPE MANNER]]
   ```

2. aorist verbal forms with root duplications and sense negation, functioning as temporal adverbs. For instance for the non-lexicalized collocation *yapar yapmaz*, where items have the parses

   ```
   [[CAT VERB] [ROOT yap] [SENSE POS]
       [TAM1 AORIST ] [AGR 3SG]]
   ```

   ```
   [[CAT VERB] [ROOT yap] [SENSE NEG]
       [TAM1 AORIST ] [AGR 3SG]]
   ```

   respectively, the preprocessor generates the feature sequence

   ```
   [[CAT VERB] [ROOT koS] [SENSE POS]
       [TAM1 AORIST] [AGR 3SG]
       [CONV ADVERB DUP-AOR] [TYPE TEMP]]
   ```

3. duplicated verbal and derived adverbial forms with the same verbal root acting as temporal adverbs, e.g., *gitti gideli*,

4. emphatic adjectival forms involving duplication and question clitic, e.g., *güzel mi güzel* (beautiful question-clitic beautiful– very beautiful)

5. adjective or noun duplications that act as manner adverbs, e.g., *hızlı hızlı*, *ev ev*,

This module recognizes all such forms and coalesces them into new feature structures reflecting the final structure along with any inflectional information.

- The preprocessor then converts each parse into a hierarchical feature structure so that the inflectional feature of the form with the last category conversion (if any) are at the top level. Thus in the example above for *geldiğimdeki*, the following feature structure is generated:

   ```
   [[CAT VERB] [ROOT gel] [SENSE POS]
   [CONV NOUN DIK] [AGR 3SG]
   [POSS 1SG] [CASE LOC]
   [CONV ADJ REL]]
   ```

   $$\Downarrow$$

   $$\begin{bmatrix} \text{CAT} & \text{ADJ} & & \\ & \begin{bmatrix} \text{CAT} & \text{NOUN} \\ \text{AGR} & \text{3SG} \\ \text{POSS} & \text{1SG} \\ \text{CASE} & \text{LOC} \end{bmatrix} & \\ \text{STEM} & & \begin{bmatrix} \text{CAT} & \text{VERB} \\ \text{STEM} & \begin{bmatrix} \text{ROOT} & \text{gel} \\ \text{SENSE} & \text{POS} \end{bmatrix} \end{bmatrix} \\ & \text{SUFFIX} & \text{DIK} \\ \text{SUFFIX} & \text{REL} & \end{bmatrix}$$

- Finally, each such feature structure is then *projected* on a subset of its features. The features selected are

   - inflectional and certain derivational markers, and stems for open class of words,

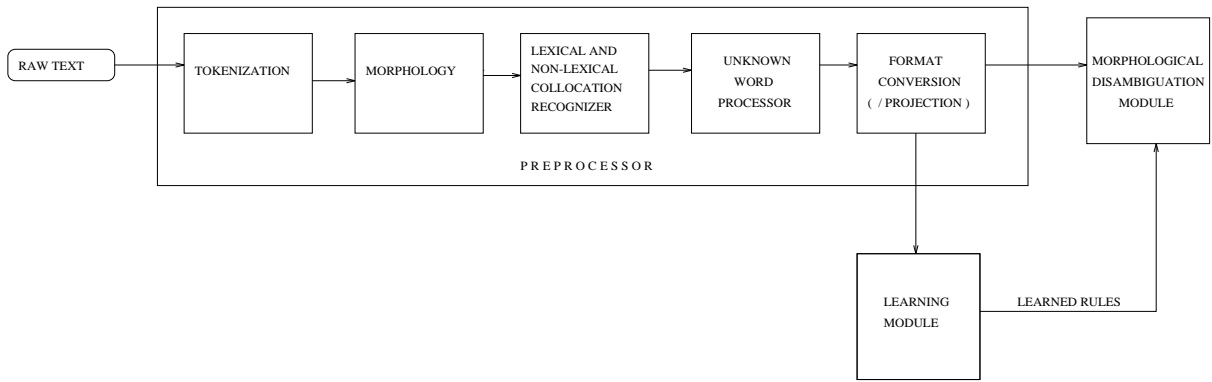

Figure 1: The structure of the constraint-based morphological disambiguation system.

– roots and certain relevant features such as subcategorization requirements for closed classes of words such as connectives, postpositions, etc.

The set of features selected for each part-of-speech category is determined by a template and hence is controllable, permitting experimentation with differing levels of information. The information selected for stems are determined by the category of the stem itself recursively.

Under certain circumstances where a token has two or more parses that agree in the selected features, those parses will be represented by a single projected parse, hence the number of parses in the (projected) training corpus may be smaller than the number of parses in the original corpus. For example, the feature structure above is projected into a feature structure such as:

$$\begin{bmatrix} \text{CAT} & \text{ADJ} \\ \text{STEM} & \begin{bmatrix} \text{CAT} & \text{NOUN} \\ \text{AGR} & \text{3SG} \\ \text{POSS} & \text{1SG} \\ \text{CASE} & \text{LOC} \\ \text{STEM} & [\text{CAT} \quad \text{VERB}] \\ \text{SUFFIX} & \text{DIK} \end{bmatrix} \\ \text{SUFFIX} & \text{REL} \end{bmatrix}$$

### 3.2 Unknown Words

Although the coverage of our morphological analyzer for Turkish (Oflazer, 1993), with about 30,000 root words and about 35,000 proper names, is very satisfactory, it is inevitable that there will be forms in the corpora being processed that are not recognized by the morphological analyzer. These are almost always foreign proper names, words adapted into the language and not in the lexicon, or very obscure technical words. These are nevertheless inflected (using Turkish word formation paradigms) with inflectional features demanded by the syntactic context and sometimes even go through derivational processes. For improved disambiguation, one has to at least recover any morphological features even if the root word is unknown. To deal with this, we have made the assumption that all unknown words have nominal roots, and built a second morphological analyzer whose (nominal) root lexicon recognizes $S^+$ where $S$ is the Turkish surface alphabet (in the two-level morphology sense), but then tries to interpret an arbitrary postfix of the unknown word as a sequence of Turkish suffixes subject to all morphographemic constraints. For instance when a form such as *talkshowumun* is entered, this second analyzer hypothesizes the following analyses:

1. [[CAT NOUN] [ROOT talkshowumun]
    [AGR 3SG] [POSS NONE] [CASE NOM]]

2. [[CAT NOUN] [ROOT talkshowumu]
    [AGR 3SG] [POSS 2SG] [CASE NOM]]

3. [[CAT NOUN] [ROOT talkshowum]
    [AGR 3SG] [POSS NONE] [CASE GEN]]

4. [[CAT NOUN] [ROOT talkshowum]
    [AGR 3SG] [POSS 2SG] [CASE NOM]]

5. [[CAT NOUN] [ROOT talkshowu]
    [AGR 3SG] [POSS 1SG] [CASE GEN]]

6. [[CAT NOUN] [ROOT talkshow]
    [AGR 3SG] [POSS 1SG] [CASE GEN]]

which are then processed just like any other during disambiguation.[8]

This however is not a sufficient solution for some very obscure situations where for the foreign word is written using its, say, English orthography, while suffixation goes on according to its English pronunciation, which may make some constraints like vowel harmony inapplicable on the graphemic representation, though harmony is in effect in the pronunciation. For instance one sees the form $Carter'a$ where the last vowel in $Carter$ is pronounced so that it harmonizes with $a$ in Turkish, while the $e$ in the surface form does not harmonize with $a$. We are nevertheless rather satisfied with our solution as in our experiments we have noted that *well below* 1% of the forms remain as unknown and these are usually item markers in formatted or itemized lists, or obscure foreign acronyms.

---

[8]Incidentally, the correct analysis is the 6[th], meaning *of my talk show*. The 5[th] one has the same morphological features except for the root.

### 3.3 Constraint Rules

The system uses rules of the sort

> if LC and RC then choose PARSE or
>
> if LC and RC then delete PARSE

where LC and RC are feature constraints on unambiguous left and right contexts of a given token, and PARSE is a feature constraint on the parse(s) that is (are) chosen (or deleted) in that context if they are subsumed by that constraint. Currently the left and right contexts can be at most 2 tokens, hence we look at a window of at most 5 tokens of which one is ambiguous. We refer to the unambiguous tokens in the context as llc (left-left context) lc (left context), rc (right context) and rrc (right-right context). Depending on the amount of unambiguous tokens in a context, our rules can have one of the following context structures, listed in order of decreasing specificity:

1.  llc, lc ____ rc, rrc

2.  llc, lc ____
    ____ rc, rrc

3.  lc ____ rc

4.  lc ____
    ____ rc

To illustrate the flavor of our rules we can give the following examples. The first example chooses parses with case feature ablative, preceding an unambiguous postposition which subcategorizes for an ablative nominal form.

```
[llc:[],lc:[],
    choose:[case:abl],
        rc:[[cat:postp,subcat:abl]],rrc:[]]
```

A second example rule is

```
  [llc:[[cat:adj,type:determiner]],
    lc:[[cat:adj,stem:[cat:noun]]],
choose:[cat:adj],
    rc:[[cat:noun,poss:'NONE']], rrc:[]].
```

which selects and adjective parse following a determiner, adjective sequence, and before a noun without a possessive marker.

Another sample rule is:

```
[llc:[],lc:[[agr:'2SG',case:gen]],
    choose:[cat:noun,poss:'2SG'],
        rc:[],rrc:[]]
```

which chooses a nominal form with a possessive marker 2SG following a pronoun with 2SG agreement and genitive case, enforcing the simplest form of noun–noun form noun phrase constraints.

Our system uses two hand-crafted sets of rules, in combination with the rules that are learned by unsupervised learning:

1. We use an initial set of hand-crafted *choose rules* to speed-up the learning process by creating disambiguated contexts over which statistics can be collected. These rules (examples of which are given above) are independent of the corpus that is to be tagged, and are linguistically motivated. They enforce some very common feature patterns especially where word order is rather strict as in NP's or PP's.[9] The motivation behind these rules is that they should improve precision without sacrificing recall. *These are rules which impose very tight constraints so as not to make any recall errors.* Our experience is that after processing with these rules, the recall is above 99% while precision improves by about 20 percentage points. *Another important feature of these rules is that they are applied even if the contexts are also ambiguous*, as the constraints are tight. That is, if each token in a sequence of, say, three ambiguous tokens have a parse matching one of the context constraints (in the proper order), then all of them are simultaneously disambiguated. In hand crafting these rules, we have used our experience from an earlier tagger (Oflazer and Kuruöz, 1994). Currently we use 288 hand-crafted choose rules.

2. We also use a set of hand-crafted heuristic *delete rules* to get rid of any very low probability parses. For instance, in Turkish, postpositions have rather strict contextual constraints and if there are tokens remaining with multiple parses one of which is a postposition reading, we delete that reading. Our experience is that these rules improve precision by about 10 to 12 additional percentage points with negligible impact on recall. Currently we use 43 hand-crafted delete rules.

### 3.4 Learning Choose Rules

Given a training corpus, with tokens annotated with possible parses (projected over selected features), we first apply the hand-crafted rules. Learning then goes on as a number of iterations over the training corpus. We proceed with the following schema which is an adaptation of Brill's formulation (Brill, 1995b):

---

[9]Turkish is a free constituent order language whose unmarked order is SOV.

1. We generate a table, called *incontext*, of all possible unambiguous contexts which contain a token with an unambiguous (projected) parse, along with a count of how many times this parse occurs unambiguously in exactly the same context in the corpus. We refer to an entry in table with a context $C$ and parse $P$ as $incontext(C, P)$.

2. We also generate a table, called *count*, of all unambiguous parses in the corpus along with a count of how many times this parse occurs in the corpus. We refer to an entry in this table with a given parse $P$, as $count(P)$.

3. We then start going over the corpus token by token generating contexts as we go.

4. For each unambiguous context encountered, $C = (\text{LC}, \text{RC})^{10}$ around an *ambiguous* token $w$ with parses $P_1, \ldots P_k$, and for each parse $P_i$, we generate a candidate rule of the sort

    if LC and RC then choose $P_i$

5. Every such candidate rule is then scored in the following fashion:

    (a) We compute
    $$P_{max} = argmax_{P_j \ (j \neq i)} \frac{count(P_i)}{count(P_j)}.$$
    $$incontext(C, P_j).$$

    (b) The score of the candidate rule is then computed as:
    $$Score_i = incontext(C, P_i) - \frac{count(P_i)}{count(P_{max})}.$$
    $$incontext(C, P_{max})$$

6. We order all candidate rules generated during one pass over the corpus, along two dimensions:

    (a) we group candidate rules by *context specificity* (given by the order in Section 3.3),

    (b) in each group, we order rules by descending *score*.

    We maintain score thresholds associated with each context specificity group: the threshold of a less specific group being higher than that of a more specific group. We then choose the top scoring rule from any group whose score equals or exceeds the threshold associated with that group. The reasoning is that we prefer more specific and/or high scoring rules: high scoring rules are applicable, in general, in more places; while more specific rules have stricter constraints and more accurate morphological parse selections, We have noted that choosing the highest scoring rule at every step may sometimes make premature commitments which can not be undone later.

---

[10] Either of LC or RC may be empty.

7. The selected rules are then applied in the matching contexts and ambiguity in those contexts is reduced. During this application the following are also performed:

    (a) if the application results in an unambiguous parse in the context of the applied rule, we increment the count associated with this parse in table *count*. We also update the *incontext* table for the same context, and other contexts which contains the disambiguated parse.

    (b) we also generate any new unambiguous contexts that this newly disambiguated token may give rise to, and add it to the *incontext* table along with count 1.

    Note that for efficiency reasons, rule candidates are not generated repeatedly during each pass over the corpus, but rather once at the beginning, and then when selected rules are applied to very specific portions of the corpus.

8. If there are no rules in any group that exceed its threshold, group thresholds are reduced by multiplying by a damping constant $d$ ($0 < d < 1$) and iterations are continued.

9. If the threshold for the most specific context falls below a given lower limit, the learning process is terminated.

Some of the rules that have been generated by this learning process are given below:

1. Disambiguate around a coordinating conjunction:

    ```
    [llc:[],lc:[],
        choose:[cat:noun,agr:3SG,case:nom],
            rc:[[cat:conn,root:ve]],
            rrc:[[cat:noun,agr:3SG,poss:NONE]]]
    ```

2. Choose participle form adjectival over a nominal reading:

    ```
    [llc:[],lc:[],
        choose:[cat:adj,suffix:yan],
            rc:[[cat:noun,agr:3SG,poss:NONE]],
            rrc:[[cat:noun,agr:3SG,poss:3SG]]].
    ```

3. Choose a nominal reading (over an adjectival) if a three token compound noun agreement can be established with the next two tokens:

    ```
    [llc:[],lc:[],
        choose:[cat:noun,agr:3SG,case:nom],
            rc:[[cat:noun,agr:3SG,poss:3SG]],
            rrc:[[cat:noun,agr:3SG,poss:3SG]]]
    ```

### 3.4.1 Contexts induced by morphological derivation

The procedure outlined in the previous section has to be modified slightly in the case when the unambiguous token in the rc position is a morphologically derived form. For such cases one has to take into consideration additional pieces of information.

We will motivate this using a simple example from Turkish. Consider the example fragment:

```
...   bir   masa+dır.
...   a     table+is
...   is    a table
```

where the first token has the morphological parses:

1. [[CAT ADJ] [ROOT bir] [TYPE CARDINAL]]
   (one)

2. [[CAT ADJ] [ROOT bir] [TYPE DETERMINER]]
   (a)

3. [[CAT ADVERB] [ROOT bir]]
   (only/merely)

and the second form has the unambiguous morphological parse:

1. [[CAT NOUN] [ROOT masa] [AGR 3SG] [POSS NONE]
      [CASE NOM] [CONV VERB NONE]
      [TAM1 PRES] [AGR 3SG]]   (is table)

which in hierarchical form corresponds to the feature structure:

$$\begin{bmatrix} \text{CAT} & \text{VERB} \\ \text{TAM1} & \text{PRES} \\ \text{AGR} & \text{3SG} \\ \text{STEM} & \begin{bmatrix} \text{CAT} & \text{NOUN} \\ \text{ROOT} & \text{masa} \\ \text{AGR} & \text{3SG} \\ \text{POSS} & \text{NONE} \\ \text{CASE} & \text{NOM} \end{bmatrix} \\ \text{SUFFIX} & \text{NONE} \end{bmatrix}$$

In the syntactic context this fragment is interpreted as

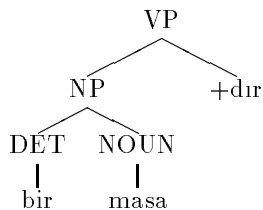

where the determiner is attached to the noun and the whole phrase is then taken as a VP although the verbal marker is on the second lexical item. If, in this case, the token *bir* is considered to neighbor a token whose top level inflectional features indicate it is a verb, it is likely that *bir* will be chosen as an adverb as it precedes a verb, whereas the correct parse is the determiner reading.

In such a case where the right context of an ambiguous token is a derived form, one has to consider as the right context, both the top level features of final form, and the *stem* from which it was derived. During the set-up of the *incontext* table, such a context is entered twice: once with the top level feature constraints of the immediate unambiguous right-context, and once with the feature constraints of the stem. The unambiguous token in the right context is also entered to the *count* table once with its top level feature structure and once with the feature structure of the stem.

When generating candidate choose or delete rules, for contexts where rc is a derived form and rrc is empty, we actually generate two candidates rules for each ambiguous token in that context:

1. if llc, lc and rc then choose/delete $P_i$.

2. if llc, lc and $stem(\text{rc})$ then choose/delete $P_i$.

These candidate rules are then evaluated as described above. In general all derivations in a lexical form have to be considered though we have noted that considering one level gives satisfactory results.

### 3.4.2 Ignoring Features

Some morphological features are only meaningful or relevant for disambiguation only when they appear to the left or to the right of the token to be disambiguated. For instance, in the case of Turkish, the CASE feature of a nominal form is only useful in the immediate left context, while the POSS (the possessive agreement marker) is useful only in the right context. If these features along with their possible values are included in context positions where they are not relevant, they "split" scores and hence cause the selection of some other irrelevant rule. Using the maxim that union gives strength, we create contexts so that features not relevant to a context position are not included, thereby treating context that differ in these features as same.[11]

### 3.5 Learning Delete Rules

For choosing delete rules we have experimented with two approaches. One obvious approach is to use the formulation described above for learning choose rules, but instead of generating choose rules, pick the parses that score (significantly) worse than and generate delete rules for such parses. We have implemented this approach and found that it is not very desirable due to two reasons:

1. it generates far too many delete rules, and

2. it impacts recall seriously without a corresponding increase in precision.

The second approach that we have used is considerably simpler. We first reprocess the training corpus but this time use a second set of projection templates, and apply initial rules, learned choose rules and heuristic delete rules. Then for every unambiguous context $C = (\text{LC}, \text{RC})$, with either an immediate left, or an immediate right components or both (so

---

[11]Obviously these features are specific to a language.

the contexts used here are the last 3 in Section 3.3), a score
$$\frac{incontext(C, P_i)}{count(P_i)}$$
for each parse $P_i$ of the (still) ambiguous token, is computed. Then, delete rules of the sort

if LC and RC then delete $P_i$

are generated for all parses with a score below a certain fraction (0.2 in our experiments) of the highest scoring parse. In this process, our main goal is to remove any seriously improbable parses which may somehow survive all the previous choose and delete constraints applied so far. Using a second set of templates which are more specific than the templates used during the learning of the choose rules, we introduce features we were originally projected out. Our experience has been that less strict contexts (e.g., just a lc or rc) generate very useful delete rules, which basically weed out what can (almost) never happen as it is certainly not very feasible to formulate hand-crafted rules that specify what sequences of features are not possible.

Some of the interesting delete rules learned here are:

1. Delete the first of two consecutive verb parses:

   ```
   [llc:[],lc:[],
       delete:[cat:verb],
           rc:[[cat:verb]],rrc:[]]
   ```

2. Delete accusative case marked noun parse before a postposition that subcategorizes for a nominative noun:

   ```
   [llc:[],lc:[],
    delete:[cat:noun,agr:3SG,poss:NONE,case:acc],
       rc:[[cat:postp,subcat:nom]],rrc:[]].
   ```

3. Delete the accusative case marked parse without any possessive marking, if the previous form has genitive case marking (signaling a genitive-possessive NP construction):

   ```
   [llc:[],
    lc:[[cat:noun,agr:3SG,poss:NONE,case:gen]],
    delete:[cat:noun,agr:3SG,poss:NONE,case:acc],
       rc:[],rrc:[]].
   ```

### 3.6 Using context statistics to delete parses

After applying hand-crafted rules to a text to be disambiguated we arrive at a state where ambiguity is about 1.10 to 1.15 parses per token (down from 1.70 to 1.80 parses per token) without any serious loss on recall. This state allows statistics to be collected over unambiguous contexts. To remove additional parses which never appear in any unambiguous context we use the scoring described above for choosing delete rules, to discard parses on the current text based on context statistics.[12] We make three passes over the current text, scoring parses in unambiguous contexts of the form used in generating delete rules, and discarding parses whose score is below a certain fraction of the maximum scoring parse, on the fly. The only difference with the scoring used for delete rules, is that the score of a parse $P_i$ here is a weighted sum of the quantity

$$\frac{incontext(C, P_i)}{count(P_i)}$$

evaluated for three contexts in the case both the lc and rc are unambiguous.

### 3.7 Steps in Disambiguating a Text

Given a new text annotated with all morphological parses (this time the parses are not projected), we proceed with the following steps for disambiguation:

1. The initial hand-crafted choose rules are applied first. These rules always constrain top level inflectional features, and hence, any stems from derivational processes are not considered unless explicitly indicated in the constraint itself.

2. The hand-crafted delete clean-up rules are applied.

3. Context statistics described in the preceding section are used to discard further parses.

4. The choose rules that have been learned earlier, are then repeatedly applied to unambiguous contexts, until no more ambiguity reduction is possible. During the application of these rules, if the immediate right context of a token is a derived form, then the stem of the right context is also checked against the constraint imposed by the rule. So if the rule right context constraint subsumes the top level feature structure or the stem feature structure, then the rule succeeds and is applied if all other constraints are also satisfied.

5. Finally, the delete rules that have been learned are applied repeatedly to unambiguous contexts, until no more ambiguity reduction is possible.

## 4 Experimental Results

We have applied our learning system to two Turkish texts. Some statistics on these texts are given in Table 1. The first text labeled ARK is a short text on near eastern archaeology. The second text from which fragments whose labels start with C are derived, is a book on early $20^{th}$ history of Turkish Republic.

In Table 1, the tokens considered are that are generated after morphological analysis, unknown word processing and any lexical coalescing is done. The words that are unknown are those that could not even be processed by the unknown noun processor. Whenever an unknown word had more than one parse it was counted under the appropriate group.

---

[12] Please note that delete rules learned may be applied to future texts to be disambiguated, while this step is applied to the current text on which disambiguation is performed.

| Text | Sentences | Tokens | Distribution of Morphological Parses | | | | | |
|---|---|---|---|---|---|---|---|---|
| | | | 0 | 1 | 2 | 3 | 4 | >4 |
| ARK | 492 | 7,928 | 0.15% | 49.34% | 30.93% | 9.19% | 8.46% | 1.93% |
| C2400 | 2,407 | 39,800 | 0.03% | 50.56% | 28.66% | 10.12% | 8.16% | 2.47% |
| C270 | 270 | 5212 | 0.02% | 50.63% | 30.68% | 8.62% | 8.36& | 1.69% |

Table 1: Statistics on Texts

We learned rules from ARK itself, and on the first 500, 1000, and 2000 sentence portions of C2400. C270 which was from the remaining 400 sentences of C2400 was set aside for testing. Gold standard disambiguated versions for ARK, C270 were prepared manually to evaluate the automatically tagged versions.

Our results are summarized in the following set of tables. Tables 2 and 3 give the ambiguity, recall and precision initially, after hand-crafted rules are applied, and after the contextual statistics are used to remove parses – all applications being cumulative. The rows labeled BASE give the initial state of the text to be tagged. The rows labeled INITIAL CHOOSE give the state after hand-crafted choose rules are applied, while the rows labeled INITIAL DELETE give the state after the hand-crafted choose and delete rules are applied. The rows labeled CONTEXT STATISTICS give the state after the rules are applied and context statistics are used (as described earlier) to remove additional parses.

| Disambiguation Stage | Ambiguity | Recall (%) | Pre. (%) |
|---|---|---|---|
| BASE | 1.828 | 100.00 | 54.69 |
| INITIAL CHOOSE | 1.339 | 99.28 | 74.13 |
| INITIAL DELETE | 1.110 | 99.08 | 88.91 |
| CONTEXT STATISTICS | 1.032 | 97.38 | 94.35 |

Table 2: Average parses, recall and precision for text ARK

| Disambiguation Stage | Ambiguity | Recall (%) | Pre. (%) |
|---|---|---|---|
| BASE | 1.719 | 100.00 | 58.18 |
| INITIAL CHOOSE | 1.353 | 99.16 | 73.27 |
| INITIAL DELETE | 1.130 | 98.73 | 87.24 |
| CONTEXT STATISTICS | 1.038 | 96.70 | 93.15 |

Table 3: Average parses, recall and precision for text C270

Tables 5 and 6 present the results of further disambiguation of ARK, and C270 using rules learned from training texts C500, C1000, C2000 and ARK. These rules are applied after the last stage in the tables above.[13] The number of rules learned are given in Table 4.[14]

| Training Text | Choose Rules | Delete Rules |
|---|---|---|
| ARK | 23 | 89 |
| C500 | 11 | 113 |
| C1000 | 29 | 195 |
| C2000 | 61 | 245 |

Table 4: Number of choose and delete rules learned from training texts.

| Disambiguation Stage | Ambiguity | Recall (%) | Pre. (%) |
|---|---|---|---|
| Training Set ARK | | | |
| LEARNED CHOOSE | 1.029 | 97.31 | 94.52 |
| LEARNED DELETE | 1.027 | 97.20 | 94.63 |
| Training Set C500 | | | |
| LEARNED CHOOSE | 1.031 | 97.30 | 94.45 |
| LEARNED DELETE | 1.028 | 97.30 | 94.61 |
| Training Set C1000 | | | |
| LEARNED CHOOSE | 1.028 | 97.29 | 94.58 |
| LEARNED DELETE | 1.026 | 97.18 | 94.68 |
| Training Set C2000 | | | |
| LEARNED CHOOSE | 1.028 | 97.24 | 94.60 |
| LEARNED DELETE | 1.025 | 97.13 | 94.71 |

Table 5: Average parses, recall and precision for text ARK after applying learned rules.

Table 7 gives some additional statistical results at the sentence level, for each of the test texts. The columns labeled UA/C and A/C give the number and percentage of the sentences that are correctly disambiguated with one parse per token, and with more than one parse for at least one token, respectively. The columns labeled 1, 2, 3, and >3 denote the number and percentage of sentences that have 1, 2, 3, and >3 tokens, with all remaining parses incorrect. It can be seen that well 60% of the sentences are correctly morphologically disambiguated with very small number of ambiguous parses remaining.

---

[13] Please note for ARK, in the first two rows, the training and the test texts are the same.

[14] Learning iterations have been stopped when the maximum rule score fell below 7.

| Text | Sentences | | | | | | | |
|------|-------|----------|----------|----------------|----------|----------|----------|----------|
|      | Total | UA/C | A/C | C (UA/C+A/C) | 1 | 2 | 3 | >3 |
| ARK  | 494 | 220 (44.53%) | 97 (19.64%) | 317 (64.17%) | 133 (26.92%) | 41 (8.30%) | 3 (0.61%) | 0 (0.00%) |
| C270 | 270 | 116 (42.96%) | 50 (18.52%) | 166 (61.48%) | 55 (20.37%) | 27 (10.00%) | 17 (6.30%) | 5 (1.85%) |

Table 7: Disambiguation results at the sentence level using rules learned from C2000.

| Disambiguation Stage | Ambiguity | Recall (%) | Pre. (%) |
|---|---|---|---|
| Training Set ARK ||||
| LEARNED CHOOSE | 1.035 | 96.64 | 93.36 |
| LEARNED DELETE | 1.029 | 96.40 | 93.71 |
| Training Set C500 ||||
| LEARNED CHOOSE | 1.035 | 96.66 | 93.32 |
| LEARNED DELETE | 1.029 | 96.40 | 93.66 |
| Training Set C1000 ||||
| LEARNED CHOOSE | 1.035 | 96.66 | 93.34 |
| LEARNED DELETE | 1.029 | 96.42 | 93.64 |
| Training Set C2000 ||||
| LEARNED CHOOSE | 1.034 | 96.64 | 93.41 |
| LEARNED DELETE | 1.030 | 96.52 | 93.70 |

Table 6: Average parses, recall and precision for text 270 after applying learned rules.

### 4.1 Discussion of Results

We can make a number of observations from our experience: Hand-crafted rules go a long way in improving precision substantially, but in a language like Turkish, one has to code rules that allow no, or only carefully controlled derivations, otherwise lots of things go massively wrong. Thus we have used very tight and conservative rules in hand-crafting. Although the additional impact of choose and rules that are induced by the unsupervised learning is not substantial, this is to be expected as the stage at which they are used is when all the "easy" work has been done and the more notorious cases remain. An important class of rules we explicitly have avoided hand crafting are rules for disambiguating around coordinating conjunctions. We have noted that while learning choose rules, the system zeroes in rather quickly on these contexts and comes up with rather successful rules for conjunctions. Similarly, the delete rules find some interesting situations which would be virtually impossible to enumerate. Although it is easy to formulate what things can go together in a context, it is rather impossible to formulate what things can not go together.

We have also attempted to learn rules directly without applying any hand-crafted rules, but this has resulted in a failure with the learning process getting stuck fairly early. This is mainly due to the lack of sufficient unambiguous contexts to bootstrap the whole disambiguation process.

From analysis of our results we have noted that trying to choose one correct parse for every token is rather ambitious (at least for Turkish). There are a number of reasons for this:

- There are genuine ambiguities. The word *o* is either a personal or a demonstrative pronoun (in addition to being a determiner). One simply can not choose among the first two using any amount of contextual information.

- A given word may be interpreted in more than one way but with the same inflectional features, or with features not inconsistent with the syntactic context. This usually happens when the root of one of the forms is a proper prefix of the root of the other one. One would need serious amounts of semantic, or statistical root word and word form preference information for resolving these. For instance, in

    koyun       sürüsü
    koyun       sürü+sü
    sheep       herd+POSS-3SG
    (sheep      herd)

    koy+un      sürü+sü
    bay+GEN     herd+POSS-3SG
    (?? bay's   herd)

    both noun phrases are syntactically possible, though the second one is obviously nonsense. It is not clear how one would disambiguate this using just contextual or syntactic information. Another similar example is:

    kurmaya              yardım      etti
    kur+ma+ya            yardım      et+ti
    construct+INF+DAT    help        make+PAST
    helped               construct   (something)

    kurmay+a             yardım      et+ti
    military-officer+DAT help        make+PAST
    helped               the         military-officer

    where again with have a similar problem. It may be possible to resolve this one using subcategorization constraints on the object of the verb *kur* assuming it is in the very near preceding context, but this may be very unlikely as Turkish allows arbitrary adjuncts between the object and the verb.

- Turkish allows sentences to consist of a number of sentences separated by commas. Hence locating a verb in the middle of a sentence is rather difficult, as certain verbal forms also have an adjectival reading, and punctuation is not very helpful as commas have many other uses.

- The distance between two constituents (of, say, a noun phrase) that have to agree in various morphosyntactic features may be arbitrar-

ily long and this causes occasional mismatches, especially if the right nominal constituent has a surface plural marker which causes a 4-way ambiguity, as in *masaları*.

```
masalarI
1. [[CAT NOUN] [ROOT masa] [AGR 3PL]
    [POSS NONE] [CASE ACC]]
   (tables accusative)

2. [[CAT NOUN] [ROOT masa] [AGR 3PL]
    [POSS 3SG] [CASE NOM]]
   (his tables)

3. [[CAT NOUN] [ROOT masa] [AGR 3PL]
    [POSS 3PL] [CASE NOM]]
   (their tables)

4. [[CAT NOUN] [ROOT masa] [AGR 3SG]
    [POSS 3PL] [CASE NOM]]
   (their table)
```

Choosing among the last three is rather problematic if the corresponding genitive form to force agreement with is outside the context.

Among these problems, the most crucial is the second one which we believe can be solved to a great extent by using root word preference statistics and word form preference statistics. We are currently working on obtaining such statistics.

## 5 Conclusions

This paper has presented a rule-based morphological disambiguation approach which combines a set of hand-crafted constraint rules and learns additional rules to choose and delete parses, from untagged text in an unsupervised manner. We have extended the rule learning and application schemes so that the impact of various morphological phenomena and features are selectively taken into account. We have applied our approach to the morphological disambiguation of Turkish, a free–constituent order language, with agglutinative morphology, exhibiting productive inflectional and derivational processes. We have also incorporated a rather sophisticated unknown form processor which extracts any relevant inflectional or derivational markers even if the root word is unknown.

Our results indicate that by combining these hand-crafted, statistical and learned information sources, we can attain a recall of 96 to 97% with a corresponding precision of 93 to 94% and ambiguity of 1.02 to 1.03 parses per token, on test texts, however the impact of the rules that are learned is not significant as hand-crafted rules do most of the easy work at the initial stages.

## 6 Acknowledgments

We would like to thank Xerox Advanced Document Systems, and Lauri Karttunen of Xerox Parc and of Rank Xerox Research Centre (Grenoble) for providing us with the two-level transducer development software on which the morphological and unknown word recognizer were implemented. This research has been supported in part by a NATO Science for Stability Grant TU–LANGUAGE.